\documentclass{article}

    \usepackage[final, nonatbib]{neurips_2024}

\usepackage[utf8]{inputenc} % allow utf-8 input
\usepackage[T1]{fontenc}    % use 8-bit T1 fonts
\usepackage{hyperref}       % hyperlinks
\usepackage{url}            % simple URL typesetting
\usepackage{booktabs}       % professional-quality tables
\usepackage{amsfonts}       % blackboard math symbols
\usepackage{nicefrac}       % compact symbols for 1/2, etc.
\usepackage{microtype}      % microtypography
\usepackage{xcolor}         % colors
\usepackage{algorithm}
\usepackage{amsmath}

\usepackage[utf8]{inputenc} % allow utf-8 input
\usepackage[T1]{fontenc}    % use 8-bit T1 fonts
\usepackage{hyperref}       % hyperlinks
\usepackage{url}            % simple URL typesetting
\usepackage{booktabs}       % professional-quality tables
\usepackage{amsfonts}       % blackboard math symbols
\usepackage{nicefrac}       % compact symbols for 1/2, etc.
\usepackage{microtype}      % microtypography
\usepackage{graphicx}
\usepackage{cleveref}
\usepackage{graphicx}
\usepackage{subcaption}
\usepackage{booktabs}
\usepackage{wrapfig}
\usepackage{wrapfig}
\usepackage{algorithm}
\usepackage{algpseudocode}
\usepackage{xcolor}

\title{HVAC-DPT: A Decision Pretrained Transformer for HVAC Control}

\author{%
  Anaïs Berkes \\
  Department of Computer Science \& Technology\\
  University of Cambridge\\
  United Kingdom\\
  \texttt{amcb6@cam.ac.uk} \\
}

\begin{document}

\maketitle

\begin{abstract}
Building operations consume approximately 40\% of global energy, with Heating, Ventilation, and Air Conditioning (HVAC) systems responsible for up to 50\% of this consumption \cite{energy, perez2008review}. As HVAC energy demands are expected to rise, optimising system efficiency is crucial for reducing future energy use and mitigating climate change  \cite{santamouris2016cooling}.
Existing control strategies lack generalisation and require extensive training and data, limiting their rapid deployment across diverse buildings. This paper introduces HVAC-DPT, a Decision-Pretrained Transformer using in-context Reinforcement Learning (RL) for multi-zone HVAC control. HVAC-DPT frames HVAC control as a sequential prediction task, training a causal transformer on interaction histories generated by diverse RL agents. This approach enables HVAC-DPT to refine its policy in-context, without modifying network parameters, allowing for deployment across different buildings without the need for additional training or data collection. HVAC-DPT reduces energy consumption in unseen buildings by 45\% compared to the baseline controller, offering a scalable and effective approach to mitigating the increasing environmental impact of HVAC systems.
\end{abstract}

\section{Introduction and related work}
\label{sec:intro_related}

Advanced controllers have the potential to significantly reduce HVAC energy consumption \cite{drgovna2020all}, but most buildings continue to rely on inefficient, rule-based systems.
Although various model-based, data-driven, and learning-based HVAC control strategies have been proposed \cite{drgovna2020all, wang2020reinforcement}, it remains a significant challenge to scale these methods across diverse building types. Model Predictive Control is limited by its reliance on precise and building-specific models, while RL requires extensive training, lasting months or years \cite{gs2023mitigating}, which often leads to suboptimal performance and occupant discomfort during the learning phase \cite{zhang2021joint}. RL also suffers from severe sample inefficiency, demanding significant amounts of sensor data  and requiring retraining for each new building. Even with transfer learning, significant data collection and customisation is still needed to address the variability in building structures and thermal dynamics between the buildings used for training and new target buildings \cite{zisman2023emergence, zhang2022diversity}.

The transformer architecture \cite{vaswani2017attention} has been widely adopted in key areas of machine learning. One major feature of transformers is in-context learning, which makes it possible for them to adapt to new tasks after extensive pretraining \cite{sinii2023context}. 
Recent research, such as the Decision-Pretrained Transformer (DPT) by Lee \textit{et al.} \cite{lee2024supervised} and Algorithm Distillation by Laskin \textit{et al.} \cite{laskin2022context}, effectively uses transformer-based in-context learning for sequential decision-making. These methods predict actions based on a query state and historical environment dynamics without the need for weight updates after the initial pretraining phase. Additionally, recent work demonstrates that transformers pretrained on diverse datasets can generalise to new RL tasks in-context, offering a promising approach for extracting generalist policies from offline RL data \cite{zhang2024decision, mukherjee2024pretraining, lin2023transformers}. Nevertheless, the application of in-context RL to HVAC control remains unexplored.

In response, we introduce HVAC-DPT, a pretrained decision transformer that uses in-context RL to optimise HVAC systems across multiple building zones without requiring prior data or additional training for new buildings. HVAC-DPT overcomes the limitations of existing control methods by enabling scalable, data-efficient, and generalisable deployment across diverse building types, removing the need for retraining and pre-deployment data collection in new environments.
In a year-long evaluation using EnergyPlus \cite{crawley2001energyplus}, HVAC-DPT reduced HVAC energy consumption by 45\% compared to baseline operations, demonstrating its transformative potential to reduce the carbon footprint of building operations.

\begin{figure}[hbt!]
    \centering
    \includegraphics[width=0.7\textwidth]{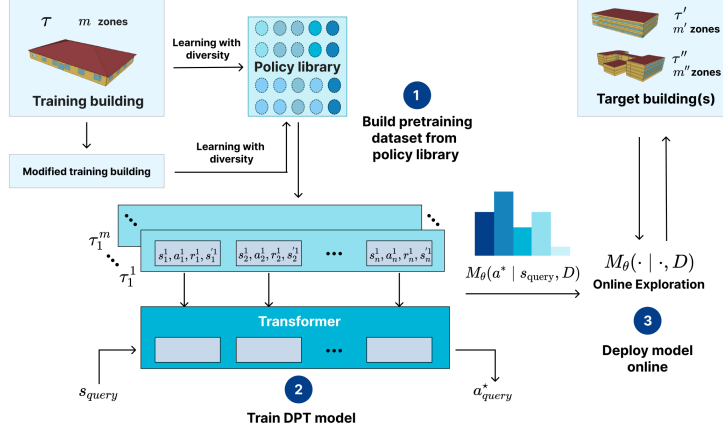}
    \caption{\label{fig:cost1} Schematic overview of the proposed methodology.}
    \label{fig:intro}
\end{figure}

\section{Problem definition}
\label{sec:problem}

\begin{wrapfigure}{r}{0.5\textwidth}
    \centering
    \captionof{table}{State variables and actions for each agent.}
    \begin{tabular}{lcl}
    \toprule
    \textbf{State} & & \textbf{Unit} \\ \midrule
    Zone mean temperature & & °C \\
    Zone mean humidity & & \% \\
    Zone occupancy & & Binary \\
    Outdoor temperature & & °C \\
    Solar radiation & & W \\
    Hour of the day & & Integer \\ \midrule
    \textbf{Action} & & \textbf{Unit} \\ \midrule
    VAV minimum damper position & & \% \\ \bottomrule
    \end{tabular}
    
    \label{tab:state_variables} 
\end{wrapfigure}

HVAC systems consist of one or more air handling units (AHUs) and variable air volume (VAV) systems, as illustrated in Figure \ref{fig:vav}. Optimising HVAC control can be framed as a sequential decision-making problem, where an agent interacts with the building, adjusts controls (e.g., VAV actuators), and receives rewards to learn control policies.

While a single agent could manage the entire building, this approach limits policy adaptability to buildings with different state-action spaces, such as those with varying numbers of VAV systems. Consequently, we model HVAC control as a multi-agent reinforcement learning (MARL) task, where each agent controls a single zone, enabling independent management across zones, similar to the approach in \cite{gs2023mitigating}.

The multi-agent Markov decision process is defined as a tuple $(N; S; A_{i}; R_{i}; T; H)$, where $N$ denotes the number of agents, $S$ is the state space, $A_i$ is the action space for agent $i$, $R : S \times A \rightarrow \Delta(R)$ is the reward function, $T : S \times A \rightarrow \Delta(S)$ is the transition function, and $H$ is the horizon.

The state space $S$, observed by all agents, includes six sensor readings, detailed in Table \ref{tab:state_variables}. Each agent's action space $A_i$ corresponds to the minimum damper position in their VAV system, ranging from 0 (closed) to 1 (fully open). The reward $R_i$ for each agent is the negative energy consumption of the VAV system during the transition from state $s$ to $s'$. The transition function $T$ is determined by EnergyPlus. Each episode has a length of $H$. Agents continuously control the VAV systems, but the control problem is modelled as episodic, with one month at 15-minute intervals constituting an episode.

\section{HVAC-DPT}
\label{sec:model}
\setlength{\textfloatsep}{10pt}
The method, illustrated in \Cref{fig:intro}, consists of three steps and builds upon the method presented by Lee \textit{et al.} \cite{lee2024supervised}: (1) A dataset $\mathcal{B}$ of RL agent interactions is collected after training a policy library of diverse RL agents in $N$ buildings. (2) A transformer model is trained to predict action labels based on a query state and the in-context dataset of interactions $D$ sampled from $\mathcal{B}$. (3) Once trained, HVAC-DPT can be deployed online in a new building by querying it for predictions of the optimal action in different states.

\begin{algorithm}[H]
\caption{HVAC-DPT}
\begin{algorithmic}[1]
    \State \textcolor{blue}{\textbf{// Dataset Generation}}
    \State Initialize empty dataset $\mathcal{B}$
    \For{$i \gets 1$ to $N$}
        \State Sample training building $\tau \sim \mathcal{T}_{\text{pre}}$
        \State Build policy library: train diverse RL policies $\pi^{i}_{\tau}$ for all zones $i \in [m]$
        \State Sample interaction dataset $D \sim \mathcal{D}_{\text{pre}}(\cdot; \tau)$ from all $\pi^{i}_{\tau}$
        \State Sample $s_{\text{query}}^{i} \sim \mathcal{D}_{\text{query}}$ and $a^\star \sim \pi^{i}_{\tau}(\cdot \mid s_{\text{query}})$
        \State Add $(s_{\text{query}}^{i}, D, a^\star)$ to $\mathcal{B}$
    \EndFor
    \State \textcolor{blue}{\textbf{// Pretraining Phase}}
    \State Initialize model $M_{\theta}$ with parameters $\theta$
    \While{not converged}
        \State Sample $(s_{\text{query}}^{i}, D, a^\star)$ from $\mathcal{B}$
        \State Predict $\hat{p}_j(\cdot) = M_{\theta}(\cdot \mid s_{\text{query}}^{i}, D_j)$
        \State Compute MSE loss with respect to $a^\star$ and backpropagate to update $\theta$
    \EndWhile
    \State \textcolor{blue}{\textbf{// Online Deployment}}
    \State Initialize $D^{i} = \{\}$ for all zones
    \State Sample target building $\tau' \sim \mathcal{T}_{\text{test}}$
    \For{ep $\gets 1$ to max\_eps}
        \For{$h \gets 1$ to $H$}
            \State $s_1 \gets \text{reset}(\tau')$
            \For{zone $i \gets 1$ to $N_{zones}^{\tau'}$}
                \State $a_h^{i} \sim M_{\theta}(\cdot \mid s_h^{i}, D^{i})$
                \State $s_{h+1}^{i}, r_h^{i} \gets \text{step}(\tau', a_h^{i})$
                \State Add $(s_1^{i}, a_1^{i}, r_1^{i}, \dots)$ to $D^{i}$
            \EndFor
        \EndFor
    \EndFor
\end{algorithmic}
\end{algorithm}

\paragraph{Dataset Generation.} The pretraining dataset $\mathcal{B}$ is collected for $N$ training buildings. HVAC-DPT generates a policy library of diverse Proximal Policy Optimisation (PPO) RL agents for the different zones in each training building $\tau$ sampled from the distribution over training buildings $\mathcal{T}_{\text{pre}}$. Both policy and environment diversity are used during training, as in \cite{gs2023mitigating}. Rollouts of these policies are used to sample an in-context dataset $D = \{s_j, a_j, s'_j, r_j\}_{j\in[n]}$ of transition tuples taken in all zones of $\tau$.

\paragraph{Pretraining.} A query state $s_{\text{query}}^{i}$ is sampled for each zone and a label $a^\star$ is sampled from an agent in the policy library. 
The in-context dataset $D$ and query state $s_{query}$ are used to train a model to predict the RL-labeled action $a^\star$ via supervised learning.
Formally,we train a GPT-2 transformer model $M$ parameterised by $\theta$, which outputs a distribution over actions $\mathcal{A}$, to minimise the expected loss over samples from the pretraining distribution:
\begin{equation}
\min_{\theta} \mathbb{E}_{P_{\text{pre}}} \sum_{j \in [n]} \ell\left(M_{\theta}(\cdot \mid s_{\text{query}}, D_j), a^\star\right).
\end{equation}
where $P_{\text{pre}}$ is the joint pretraining distribution over buildings, in-context datasets, query states and action labels.
As we have a continuous $\mathcal{A}$, we set the loss to be the Mean Squared Error (MSE).

\paragraph{Online deployment.} The model $M_{\theta}$ can be deployed online in an unseen target building $\tau'$ by initialising an empty $D^i=\{\}$ for each zone $i$ in $N_{zones}^{\tau}$. HVAC-DPT samples an action $a_h^i \sim M_{\theta}(\cdot \mid s_h^i, D^i)$ for each zone $i$ at each time-step. $D^i$ is subsequently filled with the interactions $\{s_1^i, a_1^i, r_1^i, \dots, s_H^i, a_H^i, r_H^i\}$ collected during each episode.
A key distinction to traditional RL algorithms is that there are no updates to the parameters of $M_{\theta}$. Once deployed, HVAC-DPT simply performs a computation through its forward pass to generate a distribution over actions conditioned on the in-context $D^i$ and query state $s_h^i$.

\section{Results}
\label{sec:results}

We used EnergyPlus \cite{crawley2001energyplus} and COBS \cite{zhang2020cobs} to train 100 diverse policies for $B_{train}$; further details are provided in \Cref{sec:appendix_experiment}. Four commonly used controllers were compared \cite{ahn2023alternative, gs2023mitigating}:

\begin{wrapfigure}{r}{0.55\textwidth} 
    \centering
    \vspace{-10pt} 
    \includegraphics[width=0.98\linewidth]{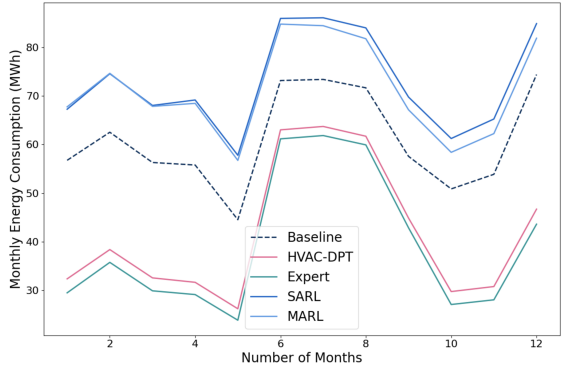}
    \caption{\label{fig:cost1} HVAC energy consumption (MWh) of different controllers during the first 12 months of deployment in building $B_{Denver}$.}
    \label{fig:results}
      \vspace{-15pt} 
\end{wrapfigure}

(1) The \textit{Baseline} controller, which maintains damper openings at 50\%; (2) The \textit{Expert} controller, implemented in the EnergyPlus model and designed specifically for each building by HVAC engineers; (3) \textit{SARL}, a single agent RL policy that controls all zones' dampers based on interaction with the target building; and (4) \textit{MARL}, which controls individual zones using the MARL framework.

\Cref{fig:results} demonstrates HVAC-DPT’s performance in $B_{Denver}$, which differs from $B_{train}$ in size and HVAC design, affecting state and action spaces. HVAC-DPT reduces energy consumption by 45\% compared to the \textit{Baseline}. 
HVAC-DPT is only 5\% less effective than the \textit{Expert} controller, despite having no prior knowledge of the building. The \textit{SARL} and \textit{MARL} controllers perform 74\% and 70\% worse, respectively, due to the extensive training required to achieve optimal performance, which can take up to 1,250 years \cite{gs2023mitigating}. More details are given in \Cref{sec:appendix_results}

\section{Conclusion}
\label{sec:conclusions}

This paper introduces HVAC-DPT, a pretrained decision transformer that uses in-context RL to optimise HVAC systems. Within the first year of deployment in new buildings, HVAC-DPT reduces energy consumption by 45\% and 70\% compared to baseline operations and RL agents respectively, all without additional training or data collection. This demonstrates HVAC-DPT's ability to generalise effectively across  buildings, addressing critical challenges in HVAC control, such as scalability, data dependency, and training efficiency.
Future work will validate HVAC-DPT in real-world settings, reinforcing its potential as a widely deployable solution for sustainable building management.

% in the face of rising HVAC energy demands.

\bibliographystyle{unsrt}
\bibliography{bib}

%\newpage
\appendix
\section{Additional system details}
\label{sec:appendix_system}

\begin{center}
\begin{figure}[hbt!]
\includegraphics[width=1.0\textwidth]{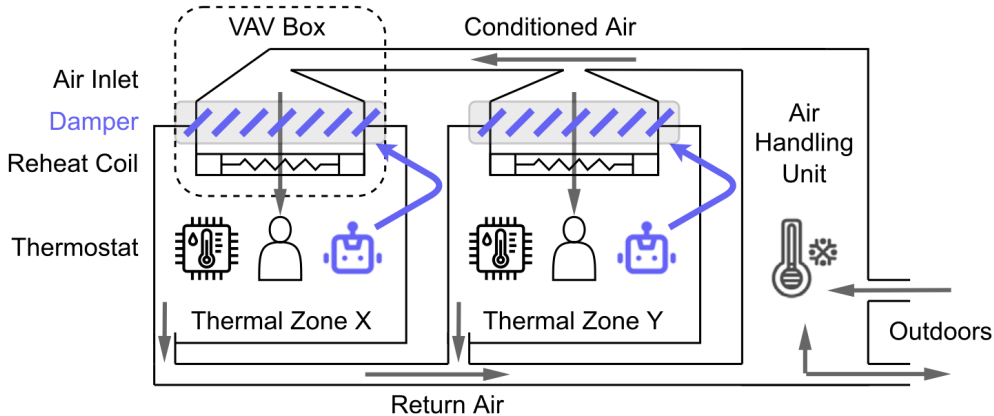}
 \caption{ Illustration of an air loop in a multi-zone building equipped with a forced-air heating and cooling system \cite{gs2023mitigating}.}
\label{fig:vav}
\end{figure}
\end{center}

 The controller in EnergyPlus adjusts the AHU and other VAV control points to ensure thermal comfort by regulating the supply air temperature and/or reheat coil power \cite{gs2023mitigating}.

\section{Additional experiment details}
\label{sec:appendix_experiment}

\paragraph{Dataset Generation} The control agents in the policy library are trained following the approach outlined in \cite{gs2023mitigating}. We use PPO with a clipping parameter $\epsilon = 0.2$, which constrains policy updates within a trust region to ensure stability. The actor and critic networks are implemented with two hidden layers, each consisting of 64 units, and use the hyperbolic tangent as the activation function. The learning rate is fixed at 0.0003, and the batch size is set to 2,976, corresponding to the length of one episode. The EnergyPlus model, used for simulating building operations, operates with 15-minute time steps, and each episode spans one month. Weather data from January 1991 is employed for training. All policies are trained using PPO within a multi-agent reinforcement learning (MARL) framework for 1,000 episodes, incorporating both environment and policy diversity as described in \cite{gs2023mitigating}. 

\paragraph{Pretraining } The HVAC-DPT model was trained using the policy library under the following conditions: a horizon of 2,967 steps, a learning rate of 0.001, and a dropout rate of 0.0. The Transformer model architecture consisted of three layers, with eight attention heads and an embedding dimension of 128. The training process was carried out for 100 trajectories over 118 epochs using the AdamW optimizer with a weight decay of 0.0001. The loss function employed was MSE. The model was evaluated using a test split of 20\%, and the training was conducted using the PyTorch framework.

\paragraph{Online Deployment}
We trained HVAC-DPT on $B_{train}$, a small office prototype building as defined by the ASHRAE Standard 90.1 \cite{ashrae2019}. $B_{train}$ is located in Denver, Colorado, and contains five thermal zones, each having an AHU and VAV system.  The total floor area of this building is 511.16 $m^2$. We used the approach presented in \cite{gs2023mitigating} to build a diverse policy library of PPO agents.
We analysed HVAC-DPT's performance on an unseen building $B_{Denver}$, a medium office prototype. $B_{Denver}$ is located in Denver, Colorado, and contains 15 thermal zones across three floors. Each floor consists of 5 zones and has an AHU and 5 VAV systems. The total floor area of this building is 4,982.19 $m^2$.
We used weather data from the year 2000 and average monthly energy consumption values over 10 runs.

\section{Additional results}
\label{sec:appendix_results}

\begin{table}[h!]
\centering
\caption{Monthly total HVAC energy consumption (in MWh) of different controllers during the first year after deployment in $B_{Denver}$. }
\vspace{0.4cm}
\renewcommand{\arraystretch}{1.5} % Increases the vertical spacing between rows
\begin{tabular}{|l|c|c|c|c|c|c|c|c|c|c|c|c|}
\hline
\textbf{Controller} & \textbf{Jan} & \textbf{Feb} & \textbf{Mar} & \textbf{Apr} & \textbf{May} & \textbf{Jun} & \textbf{Jul} & \textbf{Aug} & \textbf{Sep} & \textbf{Oct} & \textbf{Nov} & \textbf{Dec} \\ \hline
Baseline            & 56.75        & 62.52        & 56.30        & 55.78        & 44.54        & 73.15        & 73.38        & 71.65        & 57.57        & 50.87        & 53.87        & 74.38        \\ 
Expert              & 29.51        & 35.77        & 29.92        & 29.15        & 23.88        & 61.16        & 61.84        & 59.92        & 42.88        & 27.11        & 28.07        & 43.62        \\ 
MARL                & 67.74        & 74.63        & 67.84        & 68.46        & 56.73        & 84.75        & 84.42        & 81.74        & 67.13        & 58.39        & 62.24        & 81.85        \\ 
SARL                & 67.25        & 74.55        & 68.03        & 69.15        & 57.78        & 85.92        & 86.05        & 83.97        & 69.75        & 61.23        & 65.24        & 84.85        \\ 
\textbf{HVAC-DPT}   & \textbf{32.38} & \textbf{38.40} & \textbf{32.58} & \textbf{31.67} & \textbf{26.23} & \textbf{63.01} & \textbf{63.71} & \textbf{61.70} & \textbf{44.88} & \textbf{29.77} & \textbf{30.80} & \textbf{46.71} \\ \hline
\end{tabular}

\label{tab:controller_performance}
\end{table}

\begin{table}[h!]
\centering
\caption{Yearly percentile difference of total HVAC energy consumption compared to HVAC-DPT during the first year after deployment in $B_{Denver}$.}
\vspace{0.4cm} 
\renewcommand{\arraystretch}{1.5} % Increases the vertical spacing between rows
\begin{tabular}{|l|c|}
\hline
\textbf{Controller} & $\Delta_{HVAC-DPT}$  \\ \hline
Baseline            & + 45.62\% \\ 
Expert              & - 5.78\% \\ 
MARL                & + 70.56\% \\ 
SARL                & + 74.12\% \\ \hline
\end{tabular}
\label{tab:yearly_delta_to_hvac_dpt}
\end{table}

%\bibliographystyle{unsrt}
%\bibliography{bib}

\end{document}